\documentclass[journal, twoside]{IEEEtran}
\let\labelindent\relax
\usepackage{graphicx}
\usepackage{subcaption}
\usepackage{eqlist}
\usepackage{amsfonts}
\usepackage{tabularx}
\usepackage{booktabs}
\usepackage{amssymb}
\usepackage{amsmath}
\usepackage{enumitem}
\usepackage{mathtools}
\usepackage{threeparttable}
\usepackage[lined, ruled, linesnumbered, commentsnumbered]{algorithm2e}
\usepackage{lipsum}

\usepackage[usenames,dvipsnames]{xcolor}
\usepackage{comment}
\usepackage{soul}
\usepackage[clock]{ifsym}
\usepackage{flushend}

\usepackage{multirow}

\usepackage{tikz}
\usetikzlibrary{fit,shapes.misc}

\usepackage{CJKutf8}
\definecolor{bleudefrance}{rgb}{0.19, 0.55, 0.91}
\definecolor{awesome}{rgb}{1.0, 0.13, 0.32}

\definecolor{darkgreen}{rgb}{1.0, 0.0, 0.0}

\begin{document}
\title{A Dual-Arm Robot that Manipulates Heavy Plates Cooperatively with a Vacuum Lifter}
\author{Shogo Hayakawa$^{1}$, Weiwei Wan$^{1*}$, Keisuke Koyama$^1$ and Kensuke Harada$^{1,2}$
\thanks{$^{1}$Graduate School of Engineering Science, Osaka University, Japan.}%
\thanks{$^{2}$National Inst. of AIST, Japan.}%
\thanks{Contact: Weiwei Wan, {\tt\small wan@hlab.sys.es.osaka-u.ac.jp}}
}

\markboth{IEEE Automation Science and Engineering. Submission for Review, 2022}
{Hayakawa \MakeLowercase{\textit{et al.}}: A Dual-Arm Robot that Manipulates Heavy Plates Cooperatively with a Vacuum Lifter} 
\maketitle

\begin{abstract}
A vacuum lifter is widely used to hold and pick up large, heavy, and flat objects. Conventionally, when using a vacuum lifter, a human worker watches the state of a running vacuum lifter and adjusts the object's pose to maintain balance. In this work, we propose using a dual-arm robot to replace the human workers and develop planning and control methods for a dual-arm robot to raise a heavy plate with the help of a vacuum lifter. The methods help the robot determine its actions by considering the vacuum lifer's suction position and suction force limits. The essence of the methods is two-fold. First, we build a Manipulation State Graph (MSG) to store the weighted logical relations of various plate contact states and robot/vacuum lifter configurations, and search the graph to plan efficient and low-cost robot manipulation sequences. Second, we develop a velocity-based impedance controller to coordinate the robot and the vacuum lifter when lifting an object. With its help, a robot can follow the vacuum lifter's motion and realize compliant robot-vacuum lifter collaboration. The proposed planning and control methods are investigated using real-world experiments. The results show that a robot can effectively and flexibly work together with a vacuum lifter to manipulate large and heavy plate-like objects with the methods' support.
\end{abstract}
\def\abstractname{Note to Practitioners}
\begin{abstract}
This paper is motivated by the vacuum lifters used for transporting heavy plates in a factory that produces building materials. In the factory, a human worker attaches the suction cup of a vacuum lifter to a plate and controls the vacuum lifter to pull the plate up. Meanwhile, another human worker moves and lifts the plate to a goal pose following the vacuum lifter while maintaining balance. The job is dangerous as the plate is heavy and the vacuum lifter is not always strong enough to hold the plate firmly. Inspired by the usage and safety problem, we develop a planning and control method for a dual-arm robot to replace humans. The robot coordinates its motion to work with the vacuum lifter and performs lifting tasks. The vacuum lifter could remain operated by a human worker or be actuated by signals from the dual-arm robot or other third-party machines. The work is complementary to our previous study that developed planners for robots to use pulley blocks \cite{hayakawa2021}. They together provide extensive knowledge for using low-payload collaboratively robots to manipulate heavy plates.
\end{abstract}


\section{Introduction}
\label{sec:introduction}
\IEEEPARstart{C}{ollaborative} robots have been widely used in various manufacturing sites recently. They are easy to deploy and safe to replace or collaborate with human workers. Also, they have small sizes and can be adapted to many environments with high flexibility and dexterity. However, despite the advantages, collaborative robots tend to have weak payloads, making them difficult to manipulate heavy plates.

\begin{figure}[t]
    \centering
    \includegraphics[width=\linewidth]{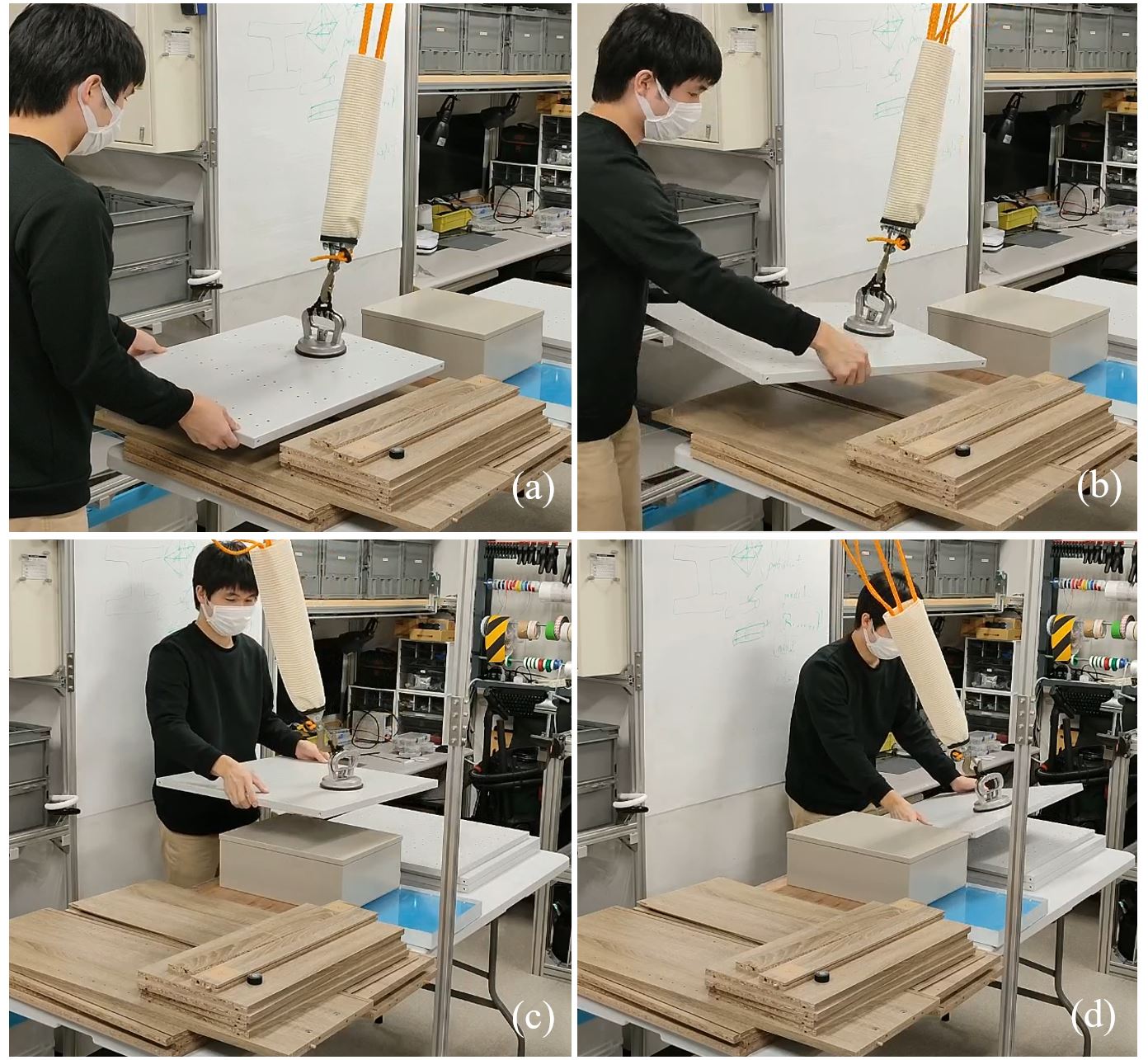}
    \caption{A human moves a heavy plate with the help of a vacuum lifter. While the lifter mostly bears the payload, the human uses his hands to maintain plate balance and control motion speed and directions. This paper studies employing a robot to replace the human and develops planning and control methods for a dual-arm robot to lift heavy and flat objects by coordinating with a vacuum lifter.}
    \label{fig:research_background}
\end{figure}

In our previous study, we solved this problem by following humans' policies and developed a planning and optimization method that enabled a dual-arm collaborative robot to use manual crane pulley blocks to lift heavy plates and flip them with non-prehensile manipulation \cite{hayakawa2021}. The method could plan flexible robotic pulling motion while avoiding collisions, reducing effort, and minimizing motor actions. Meanwhile, it could optimize a sliding pushing motion to flip target plates with a limited payload. Although the results were satisfying, the method had several shortages: The robotic manipulation was limited to a non-prehensile style since robot arms need to hold and move the pulley cables; The flipping action was prescribed as rotating the plate around an edge contact with the table; The crane hook must be attached to the center of an opposite edge using ropes to meet the analytical requirements, etc.

This paper proposes overcoming the shortages by developing planning and control methods for a dual-arm robot to cooperate with a vacuum lifter to manipulate heavy plates. Vacuum lifters are widely used in shipping and receiving areas of factories. Like pulley blocks, a vacuum lifter could assist humans in lifting and transporting heavy products. The suction cup of a vacuum lifter can be attached to any flat spot of a plate surface to provide a suction force. With the support of this suction force and careful motion planning and control, humans can manipulate plates heavier than their affordable payload, as shown in Fig. \ref{fig:research_background}. This paper takes advantage of a vacuum lifter's convenience and wide availability and develops a manipulation sequence planner and a velocity-based force controller for a dual-arm robot to lift heavy plates cooperatively with a vacuum lifter. The planner searches a Manipulation State Graph (MSG) built considering contact states, mixed prehensile and non-prehensile grasp poses, and vacuum suction forces provided by a vacuum lifter to get a flexible manipulation sequence. The controller leverages velocity-based compliant control to coordinate the motion between a robot and the vacuum lifter and lift objects to air. They together remove the need for prescribing robot actions and hooking positions and thus expand the flexibility of heavy plate manipulation by collaborative robots.

Especially, we consider the following constraints from the robot and vacuum lifter sides when developing the planning and control methods.
\setlist{nolistsep}
\begin{enumerate}[noitemsep]
    \item The vacuum lifter could either be operated by a human or actuated by a third-party system. It is independent of the robot, and there is no communication between them. The robot measures the vacuum lifter's states by forces.
    \item The attaching point of the vacuum lifter's suction cap might be far from the center of mass, making the object imbalanced. The robot needs to maintain balance during manipulation.
    \item The maximum suction force of the vacuum lifter might be smaller than the weight of a target plate. The robot needs to provide some supporting force during manipulation.
    \item Tables and surrounding obstacles might block a plate's graspable surface. The robot has to change the plate's contact state with the help of the vacuum lifter before performing prehensile actions.
    \item The dual-arm robot has a gripping hand and a pushing hand respectively on its two arms for mixed prehensile and non-prehensile manipulation.
\end{enumerate}
The developed methods comprise the following two parts. First, we build a Manipulate State Graph (MSG) to store the weighted logical relations of various plate states and robot/vacuum lifter configurations, and search the graph to plan efficient and low-cost robot manipulation sequences. Second, we develop a velocity-based impedance controller to coordinate the robot and the vacuum lifter when lifting an object. With its help, a robot can follow the vacuum lifter's motion and realize compliant robot-vacuum lifter collaboration. With our methods' support, the robot can automatically generate motion to solve the above constraints and respond flexibly to various initial plate states and vacuum lifter configurations.

In the experimental section, we study the flexibility of the proposed algorithms. The results show that our methods respond well to different plate sizes, different suction cup positions, and different initial poses. The MSG holds various combinatorics and metrics for planning flexible sequences. The compliant controller allowed the two arms to saftely move the plate following the motion of the vacuum lifter.

The remaining sections of this paper are organized as follows. Section II reviews the related work. Section III presents an overview of the used robotic systems and the algorithms. Section IV-VI jumps into the details, with graph search presented in Sections IV and V and velocity control presented in Section VI. Section VII shows the experiments and analysis. Section VIII draws conclusions and discusses future work.

\section{Related Work}
\label{sec:related_work}
We review related work from three aspects: manipulation considering contact state transitions, robots that use tools and suction cups, and multi-modal planning and control. These three aspects are the basis of our planning and control methods.

\subsection{Manipulation considering contact state transitions}
Early studies considered planning to change the state of one object concerning others for assembly tasks. For example, Hirukawa et al. \cite{hirukawa94} solved motion equations for objects with regular polyhedron shapes to obtain sticking and sliding motion. Yu et al. \cite{yu96} defined the degree of freedom of a constraint state for an object in contact. Hirukawa et al. \cite{hirukawa97}, Ji et al. \cite{ji01}, and Xiao et al. \cite{xiao01} respectively used configuration space analysis to adapt to more complicated cases. Aiyama et al. \cite{aiyama01} and Maeda et al. \cite{maeda2005planning} respectively built contact transition graphs for planning graspless manipulation. These early studies did not mention much about the robot side. More recent work further discussed robotic constraints. For example, Meeussen et al. \cite{Meeussen04} extended contact state transition by considering manipulator constraints, where an object was attached to the end flange of an industrial manipulator for controlling contact changes. Yashima et al. \cite{Yashima03} presented a probabilistic in-hand manipulation planner by considering transitions among pre-annotated contact states. Watanabe et al. \cite{watanabe06} studied the constraints of arm or finger kinematics and contact forces to automatically generate contact states and the feasible actions to transit among them. Kwak et al. \cite{kwak10} and Harada et al. \cite{harada12} respectively presented analytical methods to examine the triangles in a mesh and thus recognize contact states. Lee et al. \cite{lee2015hierarchical} proposed a hierarchical planner to find manipulation sequence considering both object-environment contact and grasp poses of free robotic hands. Aceituno-Cabezas et al. \cite{aceituno2020global} used optimization methods to generate trajectories while considering contact changes as constraints. The optimization solver determined the contact state transitions. Cheng et al. \cite{cheng2021contact} enumerated various 2D contact modes and developed a Rapidly-exploring Random Tree (RRT) based planner to generate contact transitions and motion trajectories automatically. The method was later extended to 3D cases in \cite{cheng21} by considering optimizing action forces for desired motion. 

Although the above studies exercised contact state transitions and their applications in robotics, they assumed that the objects were lighter than the maximum payload of robots, which led to limitations in task flexibility and performance. To overcome the limitations, Fakhari et al. \cite{fakhari21} developed pivoting planners that took advantage of contact state transitions and support from contact points to manipulate heavy objects. Fan et al. \cite{fan19} studied the constraints of mobile manipulators and related non-prehensile manipulation planning while taking into account contact transitions. Mohamed et al.\cite{raessa2021planning} proposed a planner to manipulate a long and heavy object using a graph-based approach. Their planner took advantage of in-hand drooping and pivoting. Zhang et al. \cite{zhang21} developed a pivoting planner for heavy objects by searching a regrasp graph of grasping poses and contact states. In this paper, we also study heavy objects but focus on planning a robot to use an external tool to assist in picking up and transferring the objects while considering contact state transitions.

\subsection{Robots that use tools and suction cups}
Previously, developing robots to pick up and use tools like humans were extensively studied, as tools are more flexible than special-purpose end-effectors and less costly than tool changers. For example, Hu et al. \cite{hu2019designing}\cite{hu2021mechanical} designed mechanical tools for two-finger parallel grippers. A robot with such grippers can use these tools to pick up various sized workpieces as well as complete screwing tasks. Murooka \cite{murooka19} developed a robot that used a hex wrench to tighten the screws on its own body. Holladay et al. \cite{holladay19} studied the force constraints in planning the tool-use motion. Toussaint et al. \cite{toussaint18} simultaneously generated the sequential manipulation and tool-use sequences by solving physical and geometrical constraints. Especially for heavy objects, Ohashi et al.\cite{ohashi2016realization} proposed using handcarts and outriggers to transport a shelf. With the tools' help, mobile robots could transport heavy a shelf beyond their maximum payload. Similarly, Scholz et al. \cite{scholz11}, and Ikeda et al. \cite{ikeda18} respectively modeled dual-arm robots that pushed hand carts and solved the related planning and control problems. Recker et al. \cite{recker2021handling} used a mobile manipulator and an external roller board as a tool to handle large and heavy construction columns. Balatti et al. \cite{balatti20} leveraged an impedance controller guided by a trajectory planner to autonomously transport and position pallet jack and relieve human workers.

In this paper, we particularly consider using vacuum lifers as tools. The functioning ends of vacuum lifters are suction cups which were widely studied in robotic manipulation research due to their flexibility. Related studies about suction cups include but are not limited to the following ones. Yamaguchi et al. \cite{yamaguchi13} installed suction mechanisms at the fingertips of robotic grippers to pick up thin objects and separate the films of plastic bags. Valencia et al. \cite{valencia20173d} used visual detection to determine an optimal suction pose. Shao et al. \cite{shao2019suction} developed a deep learning-based method to predict the suction grasp regions of objects in clutter. Cheng et al.\cite{cheng19} developed a model to estimate the state of a suction cup and used it to predict and control the suction cup for tumbling objects. Pham et al.\cite{pham2019critically} enabled manipulating boxes very fast by maximally exploiting the suction stability constraints in a time parameterized sequence. Kim et al.\cite{kim2021enhancing} attached two 3-DoF passive joints to two manipulator end flanges and connected four suction cups using a beam linkage and a central pivot joint to the passive joints for dual-arm cooperative large object transportation. Aoyagi et al.\cite{aoyagi2020bellows} designed a suction cup with force sensing ability by coating its surface with a conductive thin-film polymer and implemented adaptive grasping with multiple numbers of the design. Doi et al.\cite{Doi2020} combined a proximity sensor with a suction cup to detect deformations and realized a similar adaptivity. Huh et al.\cite{huh2021multi} developed a new type of suction cup with multiple chambers and implemented the evaluation and close-loop manipulation of objects with textured or curved surfaces. As can be seen from studies, suction cups were mostly fixed to the robot's end flanges as end-effectors. Different from them, we in this paper consider suction cups as a tool that exists externally with a robot. We assume a suction cup and its hosting vacuum lifter to be either handled and controlled by a human or a third-party machine and develop planning and control methods for a dual-arm robot to collaborate with the vacuum lifter to manipulate and lift heavy plates.

\subsection{Multi-modal planning and control}
The planning methods developed in this work are based on multi-modal motion planning \cite{hauser2010multi}\cite{suarez2018can}\cite{wan2020}\cite{baek2021pre}, which could generate motion across multiple configuration spaces identified by different grasp poses or object poses. The related studies about multi-modal motion planning were extensively reviewed in a recent work published by our group \cite{chen2021}. We are not going to repeat them here. Unlike the previous literature, we in this paper identify different configuration space modalities by considering contact states and the related grasping, pushing, and suction poses.

Our control method is based on impedance/admittance control \cite{hogan1985impedance}, which is now a standard material available in many textbooks. It is widely used to coordinate the correspondence relationship between forces and positions. Variations of impedance/admittance control include hybrid control and switches among control modes \cite{yoshikawa90}\cite{amanhoud2019dynamical}\cite{hu2021mechanical}. For synchronizing the vacuum lifter and the robot in our study, however, it is difficult to directly apply impedance/admittance control due to the difficulty of tracking the vacuum lifter's precise position. We thus use velocity-based impedance control to solve this problem \cite{chen14}. The details will be presented in Section \ref{sec:control}.

\section{Robotic Platform and Proposed Workflow}
\label{sec:platform_and_workflow}
\begin{figure}[!htbp]
    \centering
    \includegraphics[width=\linewidth]{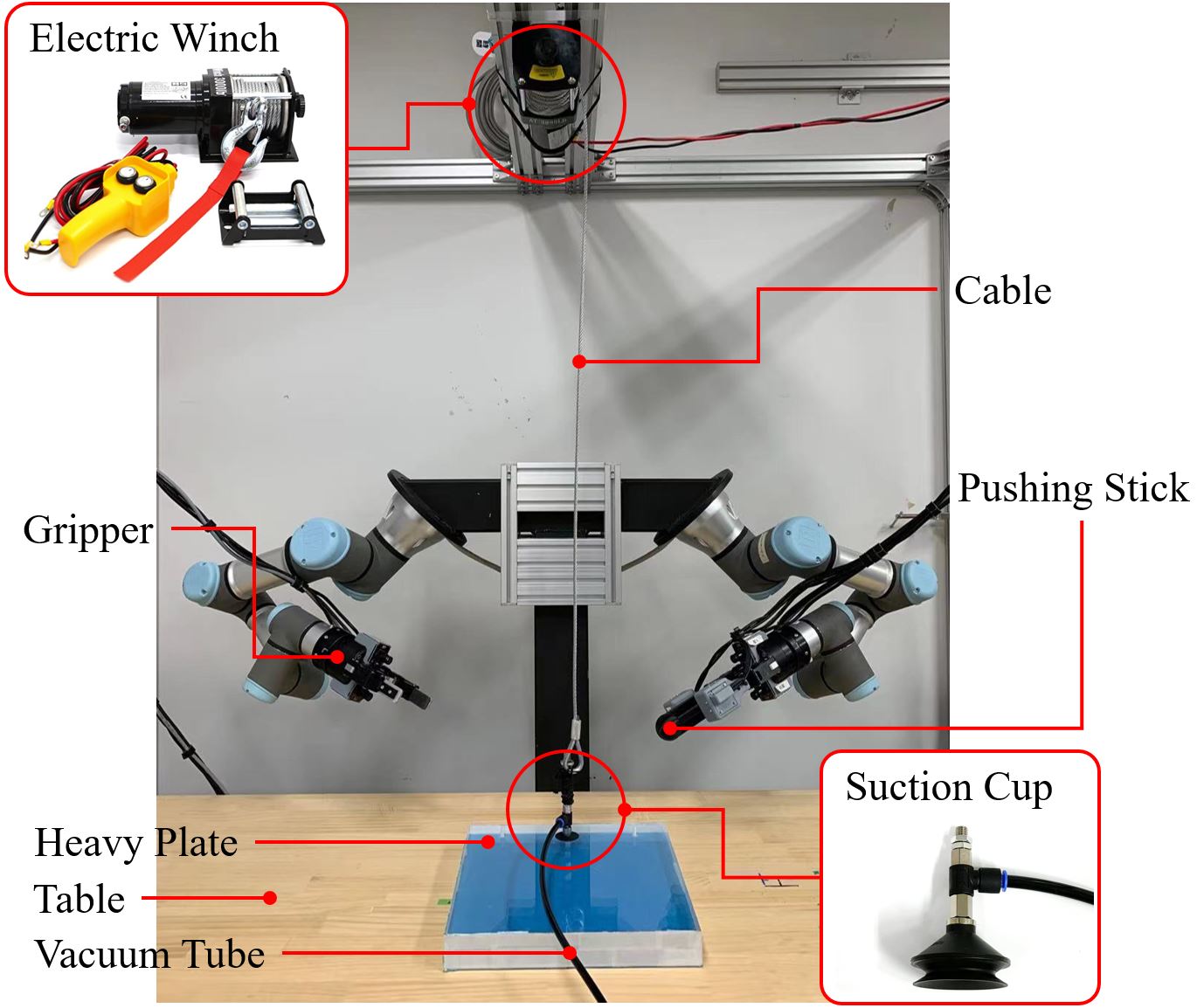}
    \caption{An exemplary dual-arm robotic platform where the proposed methods can be deployed to.}
    \label{fig:setup}
\end{figure}

Before diving into the algorithmic details, we present a concrete robotic platform and an overview of the system workflow to help conceptualize the various terminologies. Fig.\ref{fig:setup} shows the platform. Note that the proposed methods are not limited to this specific platform, but it is shown in advance for easy understanding. In the platform, an electric winch is attached above a dual-arm robot made by two UR3 robots (Universal Robots). The winch has a suction cup at its tool end to function as a vacuum lifter. A heavy and flat plate is placed in front of the robot. The suction cup is assumed to be freely attached to the plate surface by a human or a third-party machine. After attaching the suction cup, the plate can be moved by operating the winch. The suction cup is not a strong one, it could exert a suction force on the plate, but the force may not be enough to lift the plate independently. As we mentioned in the introduction section, the winch is also assumed to be operated by a human or a third-party machine, and the robot can only measure the state of the vacuum lifter using force sensing. The dual-arm robot needs to provide or find supporting forces during manipulation to make up for the insufficient suction force. The dual-arm robot has two types of end-effectors: The gripper on the left arm could be used to grasp the plate. The stick on the right arm could be used to push the plate.

The workflow of our methods includes two major sections. In the first section, we build an MSG and find a manipulation sequence that moves a plate to a non-contact state by searching the graph. In the second section, we use a velocity-based force controller to carry out leader-follower coordination and thus implement cooperative lifting up.

\begin{figure}[!htbp]
    \centering
    \includegraphics[width=.95\linewidth]{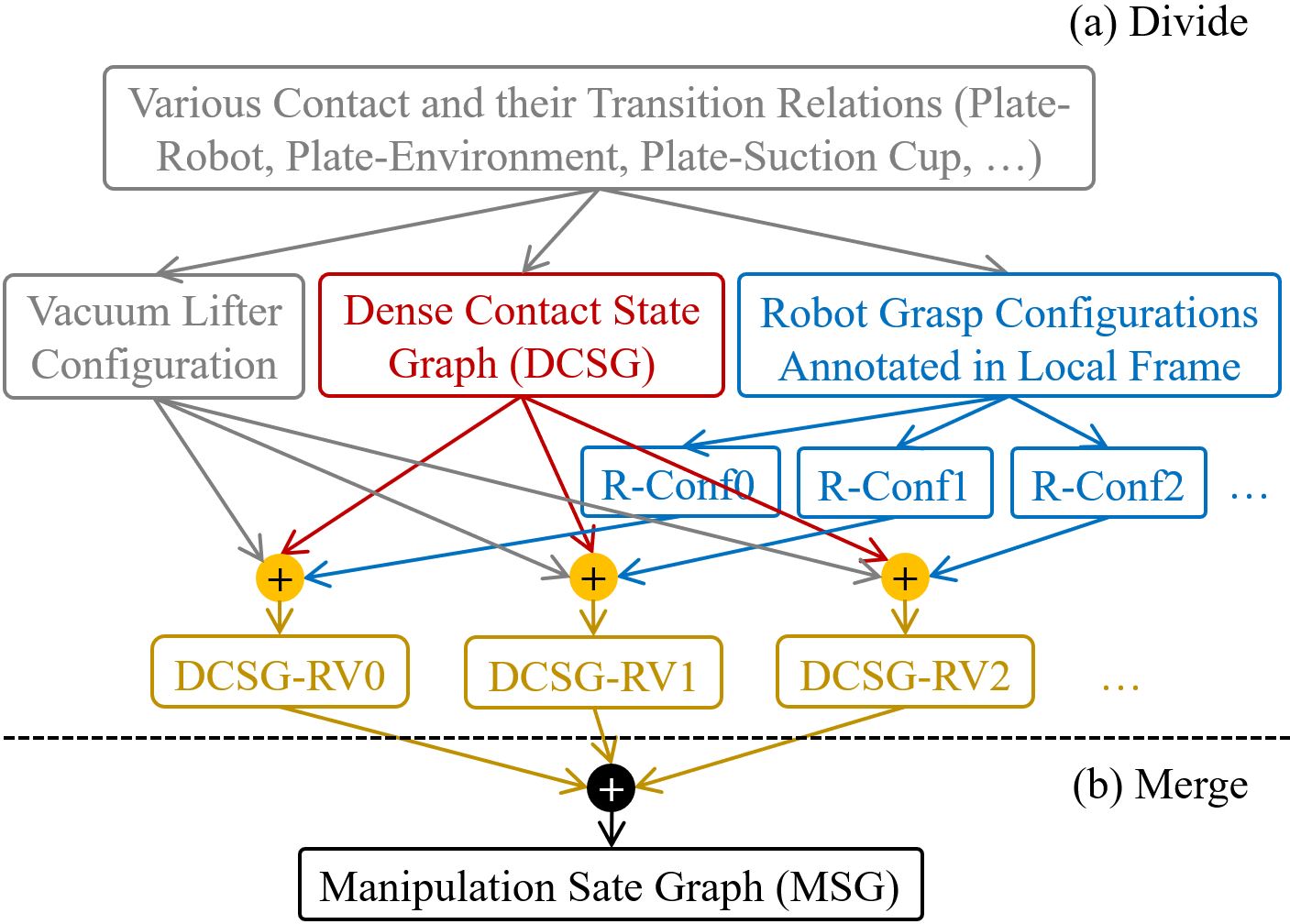}
    \caption{Two-stage divide and merge approach for creating a big contact state graph. In the division stage, the complicated contact and transition relations are divided into many DCSG-RV graphs, each identified by a pre-annotated robot-vacuum configuration. In the merging stage, the sub-graphs are combined into a big MSG considering geometrical constraints, physical constraints, and transition efforts.}
    \label{fig:divide_graph}
\end{figure}

We define a manipulation state as the contact state of a plate plus the robot hands and vacuum lifter configurations. For the specific platform described above, a manipulation state includes the plate pose, the contact of the plate with the table, the grasp configuration of the right arm, the pushing configuration of the left arm, and the suction position of the vacuum lifter. The challenge of building an MSG is that there are many contact and robot-vacuum configurations, making it complicated to identify and connect them directly. To overcome the hurdle, we consider a two-stage divide and merge approach as shown in Fig. \ref{fig:divide_graph}. First, we build a Dense Contact State Graph (DCSG) by only considering the contact between a plate and the environment. The red box in the figure shows the DCSG. Meanwhile, we annotate the robot grasp and pushing configurations in the local frame of the plate. The annotation could be either performed manually or by using grasp planning algorithms \cite{wan2020planning}. The annotation results are a lists denoted by the ``R-Conf0'', ``R-Conf1'', ... in Fig. \ref{fig:divide_graph}. Each element in the list could be a single gripping, a single pushing, or a combined gripping and pushing. The annotation results will be combined with the vacuum lifer configuration and DCSG to produce a list of DCSG-RV (Dense Contact State Graph for specific Robot-Vacuum configurations). The ``DCSG-RV0'', ``DCSG-RV1'', ... in the yellow boxes of Fig. \ref{fig:divide_graph} shows the list. In the second stage, we merge the DCSG-RVs into an MSG by reasoning grasp configurations, invalidating local transitions, and building global connections. We carefully examine geometrical and physical constraints during merging to update the connectivity of the graph. At the same time, we add weights to the graph edges by considering the robots' load changes. The final output of the divide-and-merge process is a large weighted graph that stores the logical relationship of various manipulation states. It is ready for searching and planning robot manipulation sequences while considering the vacuum lifter and environment. Although not explicitly nominated, similar methods have been used in many previous studies\cite{meeussen2004}\cite{lee2015hierarchical}\cite{wan2020}. Such methods efficiently take into account contact states and build contact state graphs. One may obtain low-cost manipulation sequences to manipulate objects by solving path planning problems over the graphs. Specifically for the MSG developed in this work, we can plan a flexible, efficient, and low-energy-cost manipulation sequence to change the plate's contact with the environment.

For cooperative lifting and transportation, we use a velocity-based leader-follower force controller to synchronize the robot and the vacuum lifter and thus adjust the imbalance caused by lags between the motion of the robot and the motion of the vacuum lifter. The reason we use velocity-based force control is that we hope to follow the motion of a vacuum lifter using force signals obtained from the force and torque sensors installed at the wrists of robotic arms. A faster vacuum lifter speed will cause a sharp drop in wrist forces and vice versa. By using velocity-based leader-follower impedance control, we can determine a high hand speed to catch up with the lifter or a low speed to wait for the lifter considering the signal changes of the force and torque sensors.

Starting from the next section, we will present the algorithmic details of our planner and controller. We use several abbreviations to denote the various concepts and graphs. Some of them have been seen in this section. Table \ref{tb:abb} summarizes them for a quick reference.

\begin{table}[!h]
\centering
\caption{\label{tb:abb} Definitions of Abbreviations}
\begin{tabularx}{\linewidth}{l|X}
\toprule
 Abbreviation & Definition \\
\midrule
PC & Principle Contact; A PC essentially describes a contact state.\\
GCR & Goal Contact Relaxation graph; A GCR comprises PCs.\\
CSG & Contact State Graph; A CSG comprises GCRs. Each contact state is identified by a unique PC.\\
DCSG & Dense Contact State Graph; Densely Sampled CSG. Each PC or contact state is sampled into multiple nodes in this graph. Each node represents a plate pose variation of the PC or contact state.\\
R-Conf & Robot Grasp Configuration; It could be a single gripping, a single pushing, or a combined gripping and pushing.\\
DCSG-RV & Dense Contact State Graph for specific Robot-Vacuum configurations; A DCSG-RV is still a DCSG but links to a unique configuration of the robot and vacuum lifter.\\
MSG & Manipulation State Graph; Each node of the graph is a manipulation state defined as the contact state of the plate plus a configuration of the robot and vacuum lifter.\\
\bottomrule
\end{tabularx}
\end{table}

\section{Preparing DCSG-RVs}
\subsection{Dense Contact State Graph (DCSG)}
\label{sec:DCSG}
We build a DCSG based on a plate's contact with the environment. Our method borrows ideas from \cite{xiao01}, where the concept of PC (Principal Contact) was introduced to define the contact states between two polyhedra. PC describes the polyhedra's actual contacting elements (faces, lines, or points). It is defined in the form of $\langle x_{A}$-$y_{B} \rangle$, where $x$, $y\in\{f, e, v\}$ with $f$ denoting a face, $e$ denoting an edge, and $v$ denoting a vertex. PCs can be merged into Goal-Contact-Relaxation (GCR) graphs. A single GCR is made by a most strongly constrained PC and its gradually relaxed neighbors. The neighbor PCs are iteratively extended from the strongest PC while considering the inclusion relations defined as follows.

Given two PCs ($PC_i = \langle a^A $-$b^B\rangle$, $PC_j = \langle c^A$-$d^B\rangle$), they have inclusion relations when 1) $c^A$ is the boundary element of $a^A$ or $d^B$ is the boundary element of $b^B$; 2) $c^A$ is the boundary element of $a^A$ or $d^B$ is equal to $b^B$; 3) $c^A$ is equal to $a^A$ or $d^B$ is the boundary element of $b^B$. According to the definition, $PC_j$ will be included as a neighbor of $PC_i$ when building GCR if any of the above condition is true.

A Contact State Graph (CSG) is a graph that holds all plate-environment contact states and their relations. It is composed of a set of GCR graphs and can be created by first identifying many GCRs and then merging them.

\begin{figure}[!htbp]
    \centering
    \includegraphics[width=\linewidth]{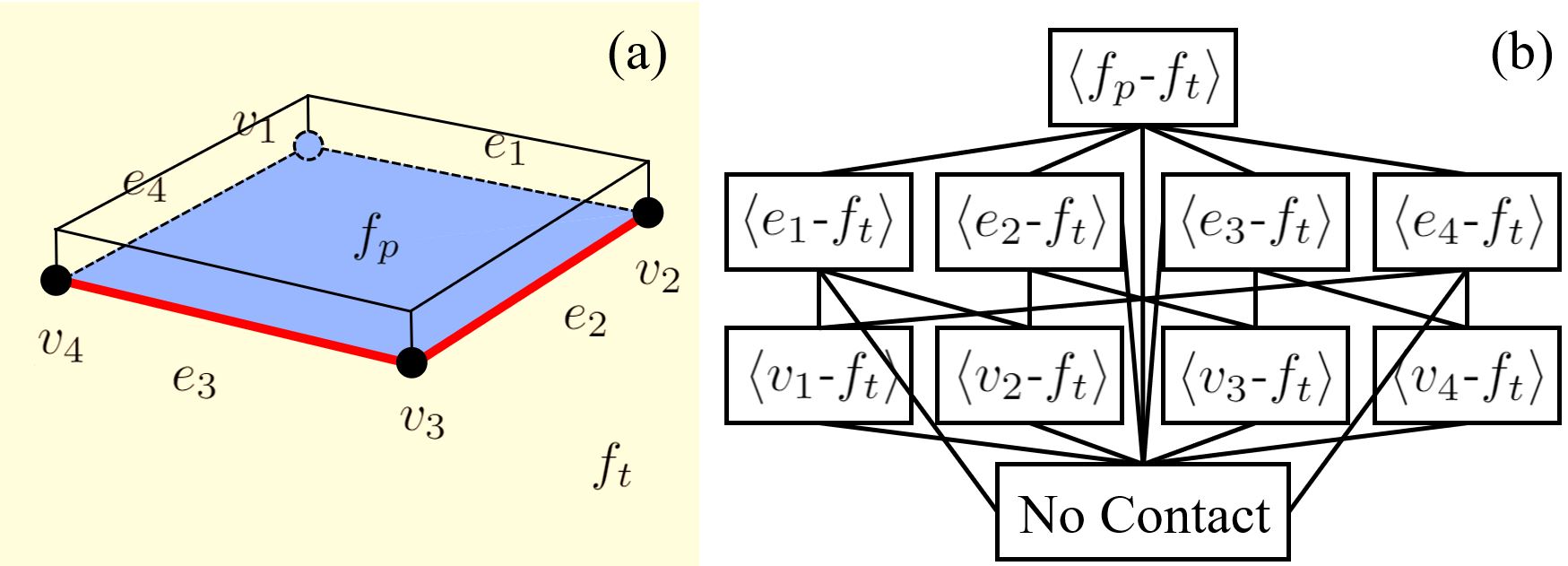}
    \caption{(a) A plate on a table. Only the elements at the bottom surface are considered and labeled. (b) PCs of the plate in (a) and the GCR formed by them.}
    \label{fig:contact_relationship}
\end{figure}

For the specific plate object and table environment assumed in this study, the above concepts could be graphically illustrated in Fig. \ref{fig:contact_relationship}(a). Here, we only consider the contact between elements at the bottom surface of the plate and the table environment. We assume the table environment is a sufficiently large flat surface so that the face is its only element. As a result, there is a single PC in which the environment most strongly constrains the plate. The PC is denoted by  $PC_0 =\langle f_p$-$f_t \rangle$ where $f_t$ is the table surface and $f_p$ is the bottom surface of the plate. A GCR can be identified by extending from $PC_0$ with PCs identified by the other elements at the bottom surface of the plate while considering the inclusion relations. Fig. \ref{fig:contact_relationship}(b) illustrates the GCR. The CSG of the plate can be created by identifying and merging all GCRs. Since we only consider the elements at the plate's bottom surface, there is only one GCR, and the CSG is essentially identical to it. Note that the ``No Contact'' node in Fig. \ref{fig:contact_relationship}(b) is a special state where the object is completely out of contact with the environment (or equally where the object is completely lifted up). Since we have not considered constraints from the robot and vacuum lifter when building the GCR and CSG, all other PCs can transit to the ``No Contact'' state.

After building the CSG, we sample each PC to make a DCSG. The sampling process increases in-state plate pose variations and thus adds flexibility to constraints from the robot and vacuum lifter. The nodes sampled from the same PC will be connected if the plate's center of mass displacement between them is smaller than a predefined threshold. The DCSG will be combined with robotic grasping and pushing configurations and also suction positions of the vacuum lifter in the next section to build DCSG-RV and finally the MSG.

\subsection{Dense Contact State Graph for specific Robot-Vacuum configurations (DCSG-RV)}
\label{sec:DCSG-RV}
\begin{figure}[!htbp]
    \centering
    \includegraphics[width=.9\linewidth]{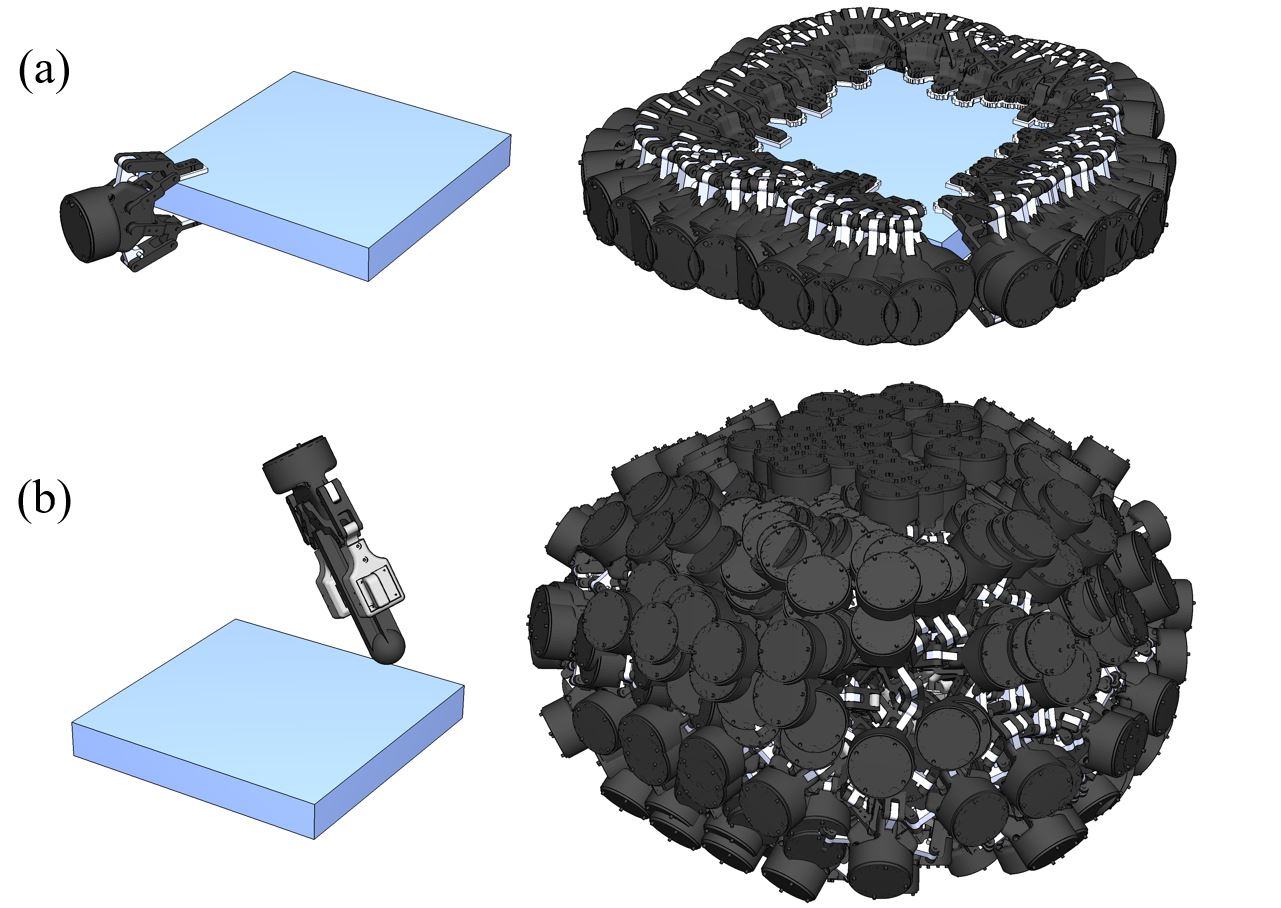}
    \caption{Pre-annotated grasping and pushing database. (a) A single grasp and the grasp database. (b) A single push and the pushing  database. The single grasp and push are shown to understand better the hand poses in the clutter.}
    \label{fig:gdb}
\end{figure}

Then, we link robot grasping/pushing configurations and a vacuum lifter's attaching configuration to the plate. The robot grasping/pushing configurations are annotated a plate's local coordinate system in advance and are used as a database in the linking process. Fig. \ref{fig:gdb} exemplifies the database. The vacuum lifter's attaching configuration is treated as an external input, for the suction cup is assumed to be attached by a human or a third-party machine. For each robot configuration in the database, we transform it to the coordinate systems of the DCSG nodes and examine the feasibility of the transformed results while considering the vacuum lifter's attaching configuration. Nodes with infeasible transformed robot configurations will be removed. The remaining nodes form a DCSG-RV graph. The transformation and feasibility check are performed per element for the robot configuration database, resulting in a collection of DCSG-RV graphs. Each graph is linked to a unique robot configuration plus the vacuum lifter's attachment.

Especially, the feasibility check examines both geometrically and physically. The geometric examination investigates robotic Inverse Kinematics (IK) and collisions with obstacles in the states represented by DCSG nodes. First, it solves the IK of the robot for the bound configuration. If IK is feasible, it detects if the robot collides with obstacles in the state represented by the DCSG nodes. A node will be kept if the robot is both IK-feasible and collision-free for the state represented by it.

The physical examination goes after the geometric examination to further investigate force feasibilities. It is carried out by solving an optimization problem using a Second-Order Cone Program (SOCP) \cite{patankar2020hand}. To formulate the optimization problem, we define $F_{g_i}$, $F_{p}$, $F_{e_j}$, $F_{s}$ as the wrenches exerted or borne by the grasping fingers, the pushing hand, the environment, and the vacuum lifter. We model $F_{g_i}$ as Soft Finger Contact with Elliptic (SFCE) approximation, model $F_{p}$ and $F_{e_j}$ as point contact with friction, and model $F_{s}$ using a conventional suction cup representation. Following the models, $F_{g_i}$, $F_{p}$, $F_{e_j}$, $F_{s}$ can be represented as
\begin{align}
    \begin{cases}
    & F_{g_i} = [f_{g_i,x}, f_{g_i,y}, f_{g_i,z}, 0, 0, \tau_{g_i,z}]^T, (i=1,2)\\
    & F_{p} = [f_{p,x}, f_{p,y}, f_{p,z}, 0, 0, 0]^T \\
    & F_{e_j} = [f_{{e_j},x}, f_{{e_j},y}, f_{{e_j},z}, 0, 0, 0]^T, (j=1,...,n)\\
    & F_{s} = [f_{s, x}, f_{s, y}, f_{s, z}, \tau_{s, x}, \tau_{s, y}, \tau_{s, z}]^T,
    \end{cases}
    \label{eq:wrenches}
\end{align}
These wrenches are described in each contact's local coordinate system $\Sigma_{g_i}$, $\Sigma_{p}$, $\Sigma_{e_j}$, and $\Sigma_{s}$, respectively.

We use a grasp map $G_\Lambda$ to represent the coordinate transformation from the contact frames to the plate's central frame $\Sigma_o$. $G_{\Lambda}$ is computed using
\begin{align}
    G_{\Lambda} = 
    \begin{bmatrix}
        {}^{o}R_{\Lambda} & 0 \\
        {}^{o}S(p_{\Lambda}) & ^{o}R_{\Lambda}
    \end{bmatrix},
    \label{eq:grasp_map}
\end{align}
where $\Lambda \in \{\Sigma_{g_i}, \Sigma_{p}, \Sigma_{e_j}, \Sigma_{s}\}$. $^{o}p_{\Lambda}$ and $^{o}R_{\Lambda}$ denote the position and the rotation matrix of $\Sigma_\Lambda$ in $\Sigma_o$. $^{o}S(p_{\Lambda})$ is a skew-symmetric matrix defined as $^{o}S(p_{\Lambda})=[^{o}p_{\Lambda}\times]$. By combining the wrenches and grasp maps in equation \eqref{eq:wrenches} and \eqref{eq:grasp_map}, we can define two large matrices as follows
\begin{align}
    \begin{cases}
        \boldsymbol{G}_{\Lambda} = [G_{g_1}, G_{g_2}, G_{p}, G_{s}, G_{e_0}, G_{e_1}, ..., G_{e_n}]\\
        \boldsymbol{F} = [F_{g_1}, F_{g_2}, F_{p}, F_{s}, F_{e_0}, F_{e_1}, ..., F_{e_n}]^T,
    \end{cases}
    \label{eq:big}
\end{align}
and represent a balanced plate state using
\begin{equation}
    \boldsymbol{G}_{\Lambda}\boldsymbol{F}+\Omega{g}=0.
    \label{eq:balance}
\end{equation}

The sub-subscript $n$ in equation \eqref{eq:big} indicates the number of point contact between a plate and the environment. For face and edge contact, we define $n$ as the number of vertices on them. In the case of a face, $n=4$. In the case of an edge, $n=2$. The notation $\Omega_{g}$ in equation \eqref{eq:balance} represents gravitational wrench, where $\Omega_{g}=[0, 0, -mg, 0,0,0]^T$ and $m$ is the mass of the plate, $g$ is the gravitational acceleration.

Using the definitions in equations \eqref{eq:wrenches}-\eqref{eq:balance}, we formulate the following optimization problem for physical examination.
\begin{subequations}\allowdisplaybreaks
    \begin{align}
        & {\small\underset{F_{g_0}, F_{g_1}, F{c}}{\text{min}}} & {\small{k_{gp}f_{gp}^++k_{\hat{s}}f_{\hat{s}}^+}} \label{eq:opgoal} \\
        & {\text{s.t.}} & \small {G}_{\Lambda}\boldsymbol{F}+\Omega{g} = 0 \label{eq:balance_sub} \\
        && 
        \begin{cases}
            \sqrt{f_{{g_i},{x}}^2 + f_{{g_i},{y}}^2 + \dfrac{\tau_{{g_i},{z}}^2}{\varepsilon_i^2}} \leq \mu_{g_i}f_{{g_i},{z}}\\
            \varepsilon_i = \dfrac{\max{{\tau_{{g_i},{z}}}^2}}{\max{{(f_{{g_i},{x}}}^2 + {f_{{g_i},{y}}}^2)}}\\
            i=1,2
        \end{cases} \label{eq:friction_suc} \\
        && \sqrt{f_{p,x}^2 + f_{p,y}^2} \leq \mu_{p}f_{p,z} \label{eq:cone_contact} \\
        && 
        \begin{cases}
            \sqrt{f_{{e_j},{x}}^2 + f_{{e_j},{y}}^2} \leq \mu_{e_j}f_{{e_j},{z}}\\
            j=1,2,...,n
        \end{cases} \label{eq:cone_env} \\
        && 
        \begin{aligned}
            \begin{cases}
                \sqrt{3}|f_{s, x}| \leq \mu_{s} f_{\hat{s}} \\
                \sqrt{3}|f_{s, y}| \leq \mu_{s} f_{\hat{s}} \\ 
                \sqrt{3}|\tau_{s, z}| \leq r\mu_{s} f_{\hat{s}}
            \end{cases}
        \end{aligned} \label{eq:cone_contact} \\
        && 
        \begin{aligned}
            \begin{cases}
                \sqrt{2}|\tau_{s, x}| \leq \pi r \kappa \\
                \sqrt{2}|\tau_{s, y}| \leq \pi r \kappa
            \end{cases}
        \end{aligned} \label{eq:material_suc} \\
        && 0 \leq f_{\hat{s}} \leq f_{\hat{s}}^+ \label{eq:suction_max}\\
        && 0 \leq f_{{g_i}, {z}}, f_{p, {z}} \leq f_{gp}^+ \label{eq:gp_max}\\
        && f_{{e_j}, {z}} \geq 0 \label{maxforce_env}
    \end{align}
    \label{eq:optproblem}
\end{subequations}
In the equation, $f_{gp}^+$ is the max force of the robot in the reversed contact normal direction. $f_{\hat{s}}^+$ is the max force of the vacuum lifter. The optimization goal is to minimize a weighted sum of the two forces. Notations $\mu_{g_i}$, $\mu_{p}$, $\mu_{e_j}$, and $\mu_{s}$ represent the friction coefficients between the plate and the grasping fingers, the pushing finger, the environment, and the suction cup of a vacuum lifter respectively. Equation (\ref{eq:friction_suc}) represents the soft finger constraint of the gripper. Equation \eqref{eq:cone_env} represents the friction cone constraints of the pushing hand and the suction cup. Equation \eqref{eq:cone_contact}-\eqref{eq:suction_max} are the constraint of a vacuum lifter's suction cup. They model a suction cup's linearized friction limit surface and elastic behavior based on the method proposed by Mahler et al. \cite{mahler2018dex}. Especially,  $f_{\hat{s}}$ equals the vacuum force caused by pressure difference minus the force that the suction cup material applies by pressing into an object. The notation $r$ in equation \eqref{eq:material_suc} indicates the radius of a suction pad. The notation $\kappa$ is a material-dependent constant of the pad. We do not differentiate the max $f_{g_i}$ and $f_p$ in equation \eqref{eq:gp_max} as we assume their max values are coherently determined by a robot's maximum payload.

We can get the optimal values of all wrenches for the contact states represented by the DSCG-RV nodes by solving this optimization problem. The obtained values will be saved as properties of the nodes for later searching and planning. On the other hand, if the problem is unsolvable for a particular node, the robot cannot provide enough force to support the contact state represented by the node, and the node is deleted from its DSCG-RV graph.

\section{Building and Searching MSG}

\subsection{MSG and edge weights}
\label{sec:complete_MSG}
\subsubsection{Merging DSCG-RVs into MSG}
In this part, we add edges across the DCSG-RVs and merge them into an MSG. We connect two nodes from two DCSG-RVs if the following two conditions are met.
\begin{itemize}
    \item The plate pose and robot configuration do not change simultaneously between the two nodes.
    \item The robot configuration linked to one DCSG-RV is a subset of the one linked to the other.
\end{itemize}
The two conditions promise a transit \cite{simeon2004manipulation} edge\footnote{In contrast, the edges inside a DCSG-RV are transfer edges. The robot moves the plate without releasing or detaching hands for a transfer edge. As a result, the plate pose changes but the grasping and pushing poses stay the same when viewing from the plate's local frame.}. When passing the edge, the robot performs regrasp to change its grasping/pushing configuration while keeping the plate stationary. The plate pose does not change, but the robot configurations have a logical implication relation from the viewpoint of set theory.

\begin{figure}[!htbp]
    \centering
    \includegraphics[width=.95\linewidth]{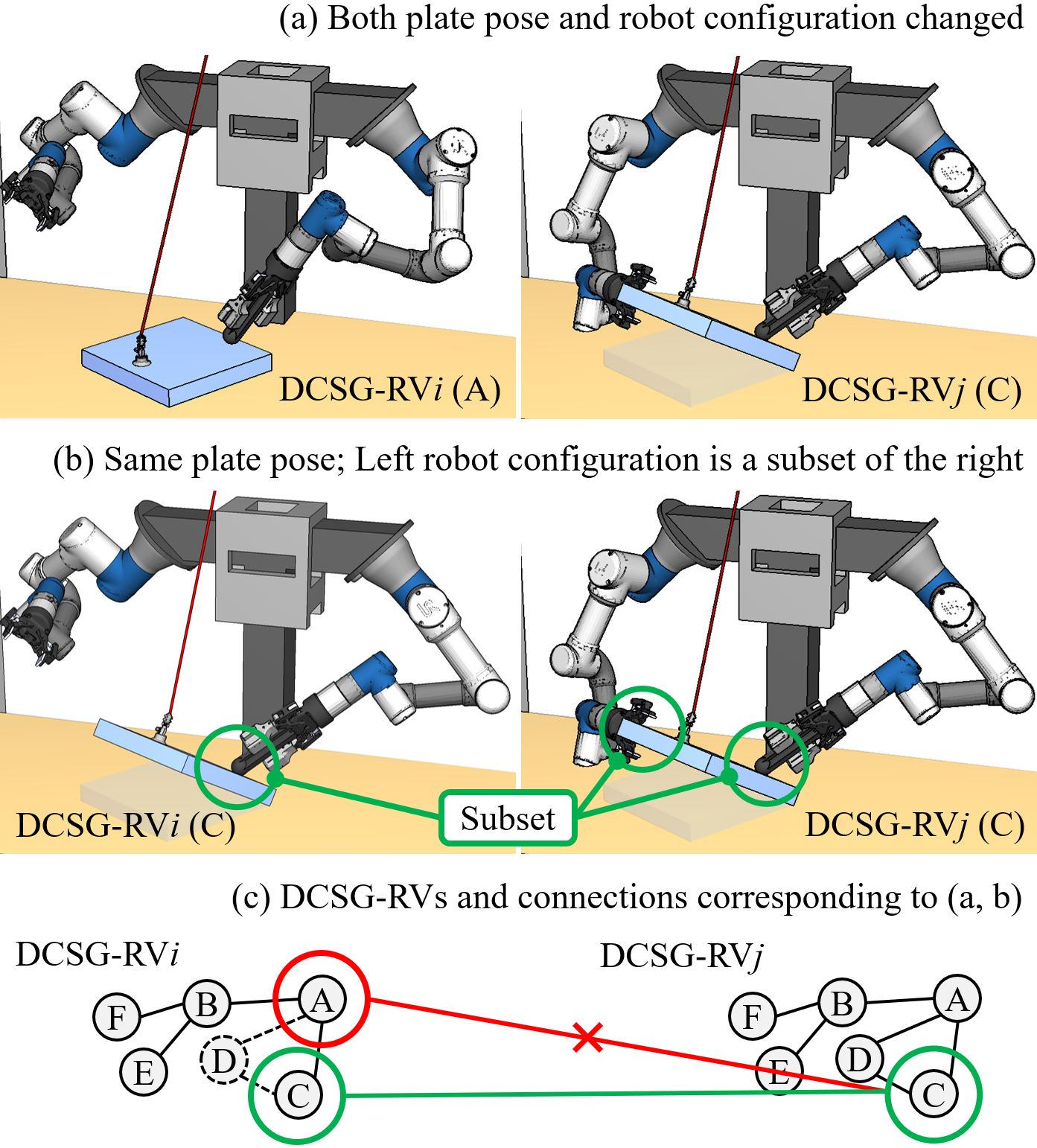}
    \caption{Merging two DCSG-RVs. The plate pose and robot configuration changed simultaneously in (a). Thus, the correspondent nodes in (c) are not connected (red edge). In contrast, the plate poses are the same and the robot configurations have a logical implication relation in (b). The correspondent nodes in (c) are consequently connected (green edge).}
    \label{fig:merge_graph}
\end{figure}

To apply the conditions, we iterate through every two-element combination of the DCSG-RVs and examine their linked robot configurations and plate poses in succession. We first check the logical relations of the robot configurations. If the two robot configurations have an implication relation, we scan the two-element combinations of correspondent nodes in the two DCSG-RVs and add a transit edge if two nodes have the same plate pose. Fig.\ref{fig:merge_graph} exemplifies an example. We can get a basic MSG after all DCSG-RV pairs are iterated and merged. 

\subsubsection{Adding weights to MSG edges}
Next, we set costs to edges of the basic MSG (both edges inside respective DCSG-RVs and edges across them) to finalize it. Primarily, we set cost values by considering the following cases.

\paragraph{Same PC, same robot configuration}
This case only appears at the inner edges of respective DCSG-RVs since the condition ``same robot configuration'' requires two nodes from the same DCSG-RV. If the nodes at the two ends of an edge have the same PC and at the same time they are affiliated with the same DCSG-RV, we set a constant value to the edge. Namely,
\begin{equation}
    cost = k \label{eq:cost_constant}
\end{equation}
This setting helps find a sequence that quickly jumps across PCs when searching the graph because the total cost becomes larger with more transitions among plate poses sampled from the same PC.
\paragraph{Different PCs, same robot configuration}
Like the first case, this case also only appears at the inner edges of respective DCSG-RVs since the robot configurations are the same. In this case, we set the cost by considering the difficulty of moving from one PC to the other. We use the restraint degree and transition difficulty measurement proposed by Yu et al. \cite{yu96} to compute the difficulty values. The method defines the following restraint matrix
\begin{align}
    T = \begin{bmatrix}
            \begin{bmatrix}
                I_3\\
                S(r_{1}) \\
            \end{bmatrix}n_{1},
            ...,
            \begin{bmatrix}
                I_3\\
                S(r_{m}) \\
            \end{bmatrix}n_{m}
        \end{bmatrix}^T \in \mathbb{R}^{m \times 6},
        \label{eq:restraint_matrix}
\end{align}
where $m$ is the number of contacts, $r_{i}, (i=1,2,...,m)$ is a vector from the object center to the contact point, $n_{i}$ is the normal vector of $i$th contact, and $I_{3}$ is the 3$\times$3 identity matrix. The contact restraint $c$ is computed as the rank of $T$.
\begin{align}
    c = \textnormal{rank}(T)
    \label{eq:restraint_degree}
\end{align}
Using the difference of contact restraints at the two end nodes, we can measure the difficulty of moving from one PC to the other. Suppose $c_{i}$ is the contact restraint of the node before the transition and $c_{j}$ is the value after the transition, we can compute the change of contact restraints by using $c_{ij} = c_{i} - c_{j}$, and measure the difficulty and thus set the cost of contact transitions as%
\begin{align}
    cost = \omega_{ij}\exp{(-c_{ij})}.
    \label{eq:transition_difficulty}
\end{align}
Here, $\omega_{ij}$ is a weight parameter. The cost makes it easier to transit to a PC with less contact restraint and helps find an efficient sequence for lifting a plate to a ``No Contact'' state.

\paragraph{Different robot configurations}
This case only appears at the additional edges used to merge the DCSG-RVs. The nodes on these edges share the same PC and plate pose since we require plate pose and robot configuration do not simultaneously change when merging the DCSG-RVs. In this case, we set the edge costs as the total joint torque that the robot needs to bear. Suppose we use ${}^{(a)}\tau_i$ as the robot joint torques in the node before transition, and ${}^{(b)}\tau_i$ as the robot joint torques in the node after the transition, the cost of the edge between the two nodes will be computed as equation (\ref{eq:cost_diffgrasp}).
\begin{align}
    cost = \sum_{i} |{}^{(a)}\tau_i| + \sum_{i} |{}^{(b)}\tau_i| \label{eq:cost_diffgrasp}
\end{align}

Note that each joint torque in the equation can be computed using $\tau = J^{T}\sum_i{F_{g_i}}$, where $J$ is the Jacobian matrix of a robot arm, and $F_{g_i}$ is the wrench obtained by solving the optimization problem in equation \eqref{eq:optproblem}. Using the total joint torque as the cost for edges that connect different robot configurations helps maintain a small effort when moving across the nodes with different robot configurations.

\subsection{Searching an MSG to find a manipulation sequence}
\begin{figure}
    \centering
    \includegraphics[width=.95\linewidth]{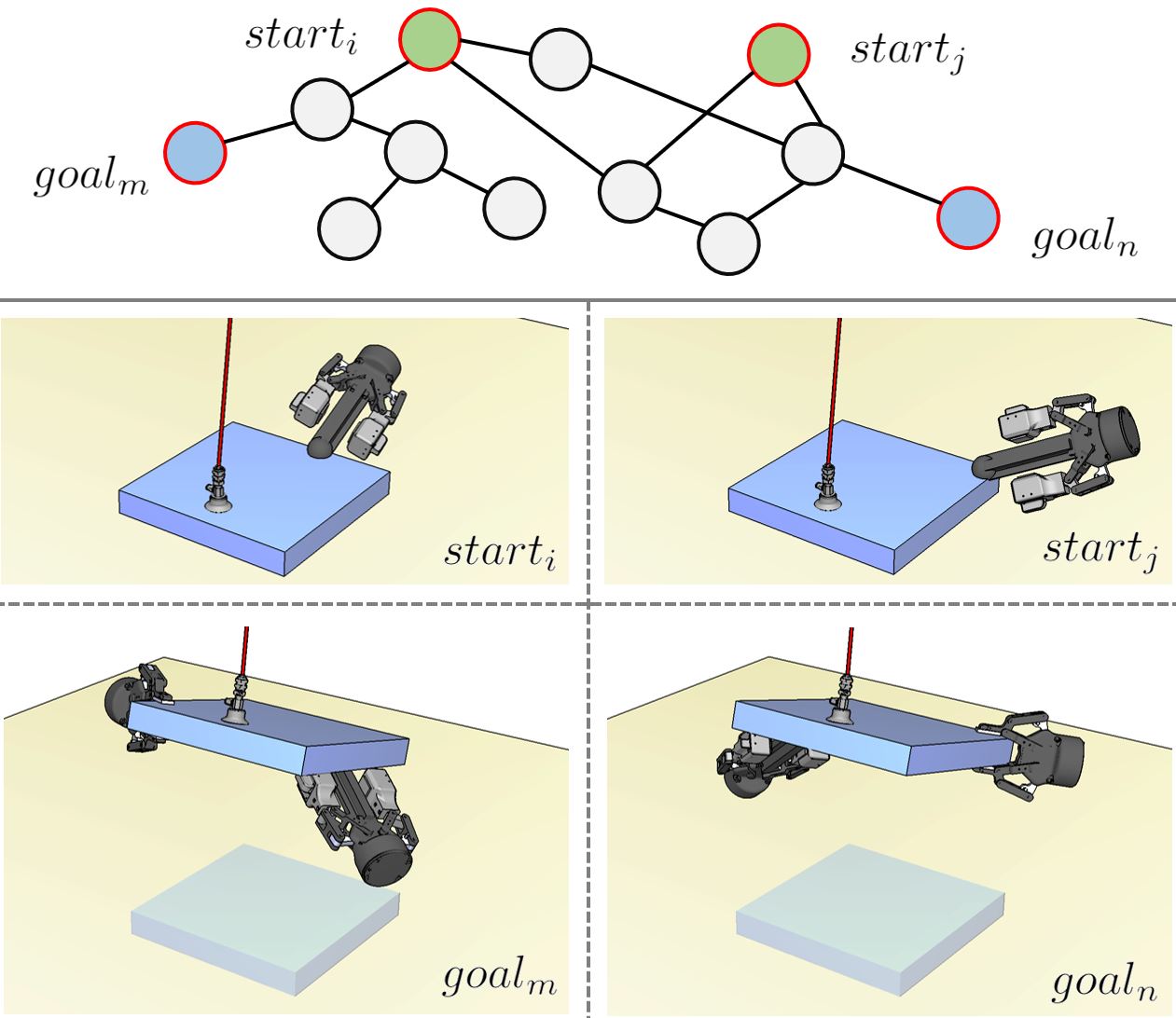}
    \caption{Searching the MSG is essentially a multi-start to multi-goal problem. The starting and goal node could be any node that implies the given initial and goal plate states. The figure exemplifies two starts and two goals. The problem is solved as long as a path between one combinatory pair is found.}
    \label{fig:find_sequence}
\end{figure}

By searching the finalized weighted MSG, we can obtain a flexible and efficient robotic manipulation sequence that lifts the plate from given initial states to goal states with the support of the vacuum lifter. Since the nodes in the MSG are both identified by contact states of the plate and robot configurations, the search problem has multiple starts and multiple goals, as illustrated in Fig.\ref{fig:find_sequence}. We solve the problem using the following routine.
\begin{enumerate}
    \item Find all nodes that imply the given initial state and the final states, and name them $start_i$, $start_{i+1}$,... and $goal_m$, $goal_{m+1}$, ... respectively.
    \item Search the sequences from one starting node to all goal nodes using the A* algorithm \cite{hart1968formal}, and save them into a list $pathlist_{j}$, where $j$ indicates that the sequences saved in it start from $start_j$.
    \item Find the sequence with the minimum total cost in $pathlist_{j}$, and define this sequence as $minpath_{j}$.
    \item Repeat step 2) for all starting nodes and get a set of $pathlist_{j}$ and hence a set of $minpath_{j}$. Save all $minpath_{j}$ as a list $min\_pathlist$.
    \item Get the sequence with the minimum force cost from $min\_pathlist$. Represent this sequence using $minpath^{*}$.
\end{enumerate}
The $minpath^{*}$ sequence is the searching result. Each node in the sequence simultaneously implies a robot configuration and a plate state. The dual-arm robot can follow the robot configurations to manipulate the plate into the states.

\section{Velocity-based Leader-Follower Control for Cooperative Lifting and Transportation}
\label{sec:control}
\begin{figure}
    \centering
    \includegraphics[width=\linewidth]{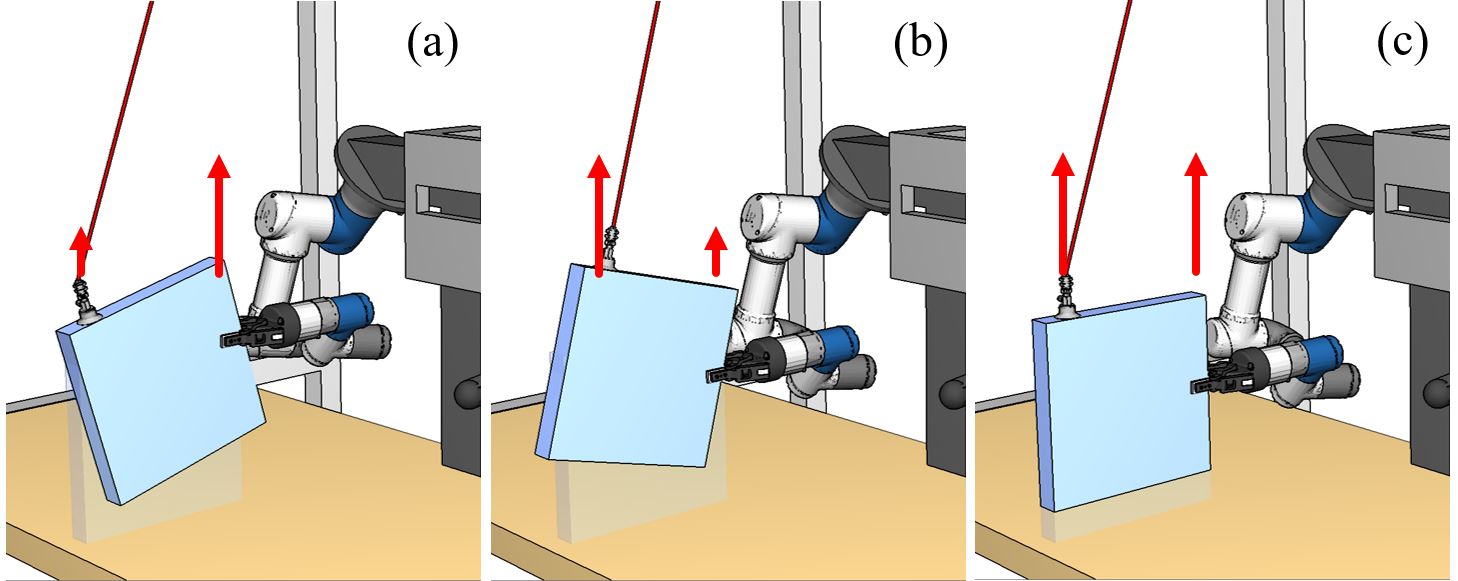}
    \caption{(a, b) Two imbalanced cases. (a) The robot arm moves faster than the vacuum lifter. (b) The vacuum lifter is faster than the robot arm. (c) A balanced collaboration.}
    \label{fig:suction_robot_balance}
\end{figure}

By specifying the goal state of the MSG searching problem to be ``No Contact'', we can find a robot sequence that manipulates the plate from an initial pose to a floating pose that is completely separated from the table and is ready to be freely moved by robot-vacuum lifter collaboration. The challenge to the robot-vacuum lifter collaboration is that there is no communication between them, and the robot has to use sensory feedback to quickly respond to the lifter motion and keep the plate balanced. Fig.\ref{fig:suction_robot_balance} demonstrates a balance problem by using the collaboration between a single robot arm and the vacuum lifter as an example. In Fig. \ref{fig:suction_robot_balance}(a), the robot arm moves faster, and the motion of the vacuum lifter is lagging behind the robot. The vacuum lifter and the robot are imbalanced. They bear a small and a large load, respectively and the robot hand is in a dangerous condition. In Fig. \ref{fig:suction_robot_balance}(b), the vacuum lifer is moving faster than the robot arm. The vacuum lifter and the robot bear a large and a small load, respectively and the vacuum lifter is also in danger. The two imbalanced cases must be avoided during the collaboration to ensure successful executions. Fig.\ref{fig:suction_robot_balance}(c) shows a balanced case where both the lifter and the robot are safe. Although the balanced plate is in a horizontal pose, its balance depends on the tension of the suction pad and the object's changing poses, and it is difficult to avoid the imbalance by simply using the positional relationship between the object and the vacuum lifter. Therefore, we focus on the velocity instead of the position and solve the balance problem by using velocity-based impedance control \cite{chen2014velocity}.

Velocity-based impedance control focuses on the relationship between the force borne by the robot and the robot hand's motion velocity. Its rule can be formulated as a variant of a conventional impedance control equation using
\begin{equation}
    M_{d}(\dot{V} - V_{d}) + B_{d}(V - V_{d}) + K_{d}\int (V - V_{d})= E,
    \label{eq:velocity-impedance}
\end{equation}
where $M_{d}$, $B_{d}$, and $K_{d}$ are the parameters of inertia, damping, and stiffness respectively, $V_{d}$ is the target value of velocity, and $E$ is the difference between the target force $F_{goal}$ and the current force $F$ and $E = F_{goal} - F$. In our implementation, we compute $F_{goal}$ following the optimization problem shown in equation \eqref{eq:opgoal} to \eqref{maxforce_env}, and measure $F$ in real time using the force and torque sensors installed at the robot wrists. By substituting the $F_{goal}$ and $F$ values into equation \eqref{eq:velocity-impedance}, we can get the velocity $V$ needed to cancel out the force difference. 

When both arms of the robot are used, simply implementing two asynchronous and respective velocity-based force controllers mentioned above will not coordinate them, for a faster speed on a single arm may lead to a substantial payload on the other one and thus result in an overload emergency. Thus, instead of respective control, we move the two arms together by asking them to follow the force changes of the pushing hand simultaneously. The choice is reasonable since
\begin{enumerate}
    \item Equation \eqref{eq:optproblem} optimized the forces to make them evenly distributed. The force changes caused by vacuum lifting will be equal under the even distribution.
    \item The pushing hand is more sensitive than force sensors than the gripper as it is non-prehensile and separates from objects easily.
\end{enumerate}
In the experimental section, we will show the results of dual-arm control and analyze the force changes in detail.

\section{Experiments and Analysis}
\label{sec:ea}
The proposed method is implemented and examined using the robot system shown in Fig.\ref{fig:setup}. The system comprises two UR3e robots with two Robotiq F85 two-finger grippers and two Robotiq FT300 force and torque sensors installed at the robotic flanges. The computer used for computation and control is a PC with Intel Core i9-9900K CPU, 32.0GB memory, and GeForce GTX 1080Ti GPU. The programming language for implementing the system is Python 3.

\begin{figure}[tbp]
    \centering
    \includegraphics[width=\linewidth]{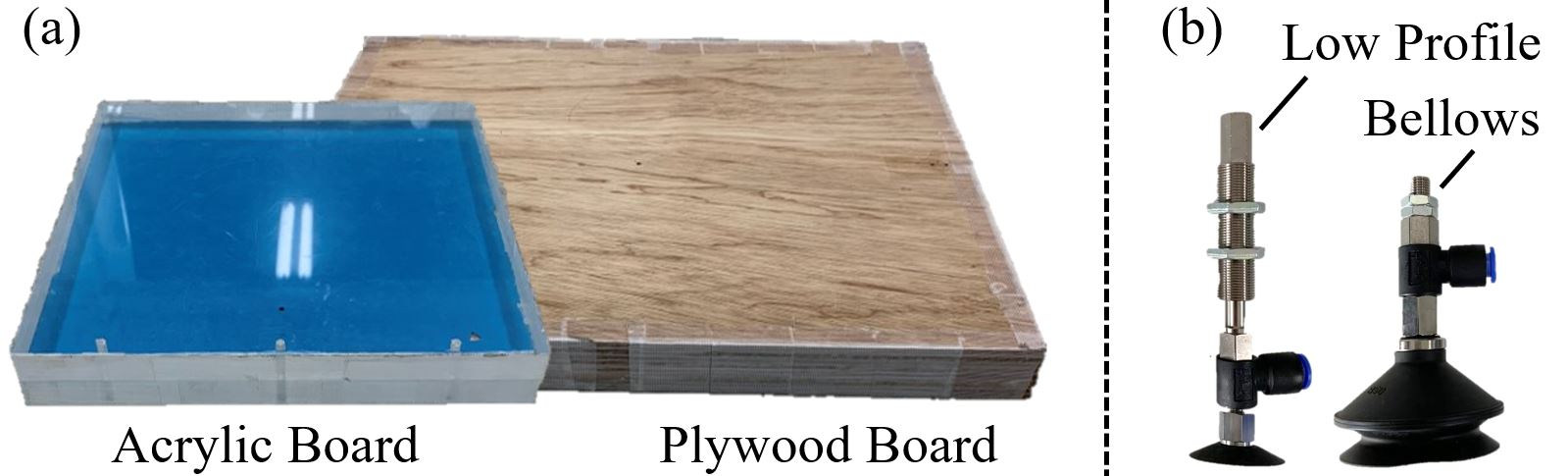}
    \caption{(a) Two different plates. (b) Two different types of suction cups.}
    \label{fig:objects}
\end{figure}

We study the performance of the proposed method by planning the dual-arm robot to lift two different plates from initial states on the table. To understand our planner's flexibility, efficiency, and efficacy, we examine and compare the planned sequence under different suction cup types, suction positions for the vacuum lifter, initial plate poses, and physical parameters. The specifications of the two different plates are shown in Table \ref{tab:objects}. Their pictures are available in Fig. \ref{fig:objects}(a). The suction cups of the vacuum lifter used in the experiments include a low profile type (diameter 3.5mm) and a bellows-type with 1.5 fold (diameter 5.5mm). Their images are available in Fig. \ref{fig:objects}(b). The maximum vacuum forces of the two suction cups under the vacuum provided by our compressor (SLP-07EED, ANEST IWATANI Corporation) and vacuum ejector (TVR-2-S10HS, TKY Co., Ltd.) are 20N and 60N, respectively.

\begin{table}[!htbp]
    \renewcommand{\arraystretch}{1.2}
    \caption{Parameters of the plates used in experiments}
    \centering
    \begin{threeparttable}
    \begin{tabular}{@{\extracolsep{0pt}}lllllll} \toprule
        Plate Name & $h$, $l$, $w$ (mm) & $m$ (kg) & Center of Mass (CoM) \\ \midrule
         Acrylic Board (AB) & 300, 300, 40 & 4.0 & Geometric Center \\
         Plywood Board (PB) & 500, 400, 44 & 6.4 & Geometric Center \\ \bottomrule
    \end{tabular} 
    \begin{tablenotes}[flushleft]
    \item (Notes) $l$ - length; $w$ - width; $h$ - height; $m$ - mass.
    \end{tablenotes}
    \end{threeparttable}
    \label{tab:objects}
\end{table}

\begin{figure}[tbp]
    \centering
    \includegraphics[width=\linewidth]{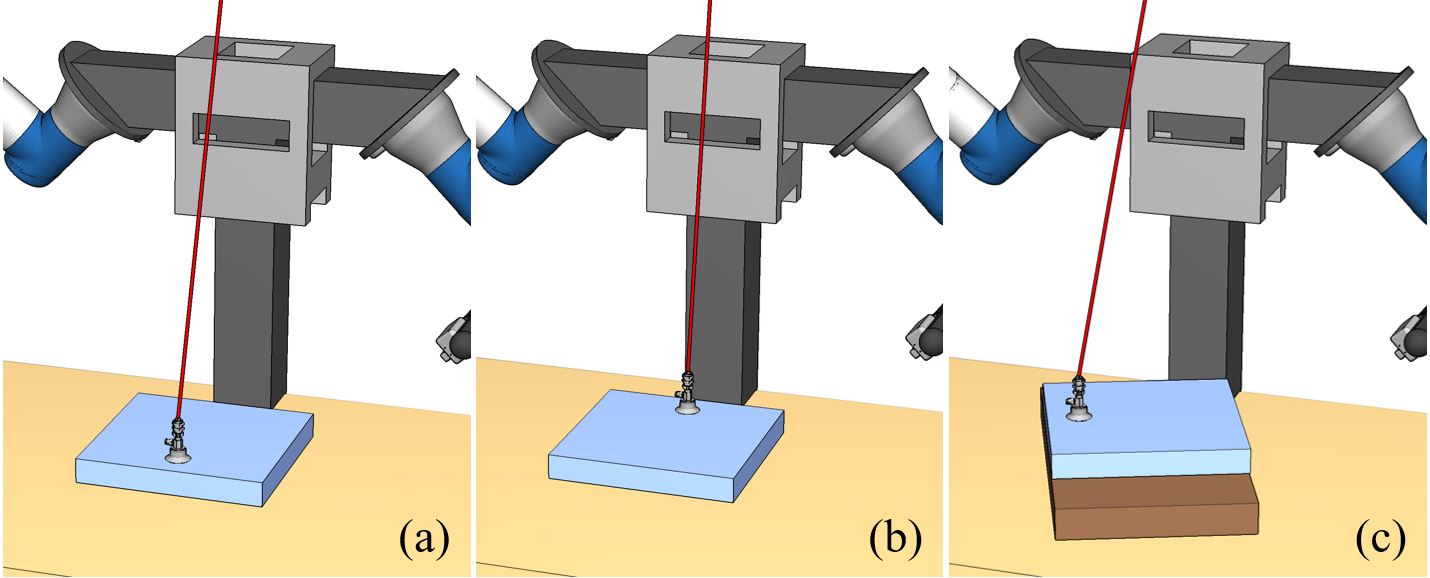}
    \caption{Three different initial states for the Acrylic Board. (a) Normal pose with front suction. (b) Normal pose with back suction. (c) Stacked pose with side suction. For the Plywood Board, we assume two initial states similar to (a) and (b).}
    \label{fig:exp_ab}
\end{figure}

\begin{figure*}[!htbp]
    \centering
    \includegraphics[width=\linewidth]{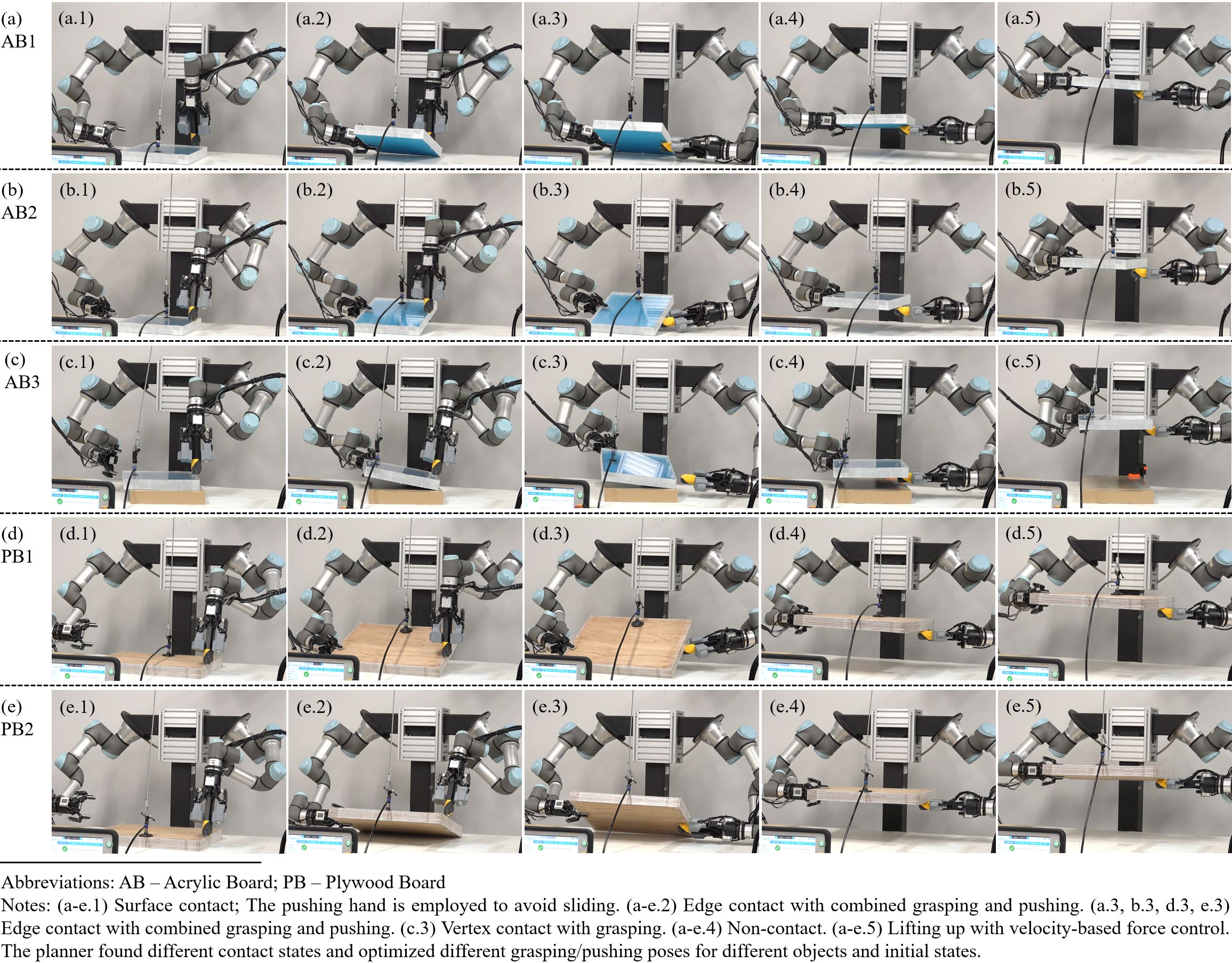}
    \caption{Planned combined prehensile and non-prehensile manipulation sequences for the Acrylic Board and the Plywood Board. The results that correspond to the three initial states of the Acrylic Board are shown in  (a-c). The results that correspond to the two initial states of the Plywood Board are shown in (d,e).}
    \label{fig:all_results}
\end{figure*}

We assumed three different initial states for the Acrylic Board, as shown in Fig. \ref{fig:exp_ab}. In the state shown in Fig. \ref{fig:exp_ab}(a, b), the plate was placed horizontally on the table in front of the dual-arm robot without rotation around the vertical axis. The vacuum lifter's suction cups were attached to the front and back area of the plate's top surface to examine how the planned results changed. In the state shown in Fig. \ref{fig:exp_ab}(c), the Acrylic Board was stacked on another wooden board and was rotated 20 degrees around the vertical axis to provide a different initial pose. The vacuum lifter's suction cup used in the experiments was the low-profile type. Its maximum vacuum force was not enough to independently lift the Acrylic Board. 

For the Plywood Board, we assumed two initial states the same as the first and second states of the Acrylic Board, except that the bellows suction cup was used instead of the low-profile one. The reason was that the low-profile suction cup could not provide enough force\footnote{Although the low-profile suction cup can provide at maximum 20N force, one should not use this value to determine if the cup could provide enough force for the robot. When a board is tilted, the suction changes from a horizontal one to a vertical one, and the maximum force provided by the cup decreases to $\mu$20N where $\mu$ is the Coulomb friction coefficient.} for the plywood board and our proposed planner failed to find a successful motion sequence.

The results for the Acrylic Board are shown in Fig. \ref{fig:all_results}(a-c). Each subfigure represents the result of a different initial state. The ``AB'' in the names is the abbreviation of Acrylic Board. The numbers 1, 2, and 3 sequentially indicate that the subfigures assumed the (a), (b), and (c) initial states in Fig. \ref{fig:exp_ab}. Full video clips of the sequences are available in the supplementary file for examining the motion details. The results include two stages. In the first stage, the robot performs combined prehensile and non-prehensile manipulation to change the board state from surface contact to a non-contact state in the air. The first four columns of Fig. \ref{fig:all_results}(a-c) show key motion frames for the first stages of the three initial states, respectively. In the second stage, the two arms follow the vacuum lifter to lift the board. The fifth column of Fig. \ref{fig:all_results}(a-c) shows the finally lifted states. From the first stage results, we could see that the planner found different ways for the robot to manipulate the board to a non-contact state. In (a) and (b), the changes of contact states were similar, but the robot chose different grasping and pushing poses to optimize the workload. The different poses were symmetric around the line connecting the plate's center of mass and the suction position since the two initial states were mirrored. In (c), the board was changed to a vertex contact state before finally being lifted to the air. The reason could be the robotic kinematic constraint, constraints from the workload optimization, or the weights along the graph edges. 

Fig. \ref{fig:all_results}(d, e) show the sequences of the Plywood Board. Like the Acrylic Board, each subfigure represents the result of a different initial state. The ``PB'' in the names is the abbreviation of Plywood Board. The numbers 1 and 2 respectively indicate that the subfigures assumed the (a) and (b) initial states in Fig. \ref{fig:exp_ab}. The video clips for the sequences are also available in the supplementary file. There is no significant difference between the planned sequence compared with the Acrylic Board. The sequences are also divided into two stages, with the first four columns showing key motions for combined prehensile and non-prehensile manipulation and the last column showing lifting control. The robots manipulated the boards to a similar edge contact state before finally raising them to the air. The only change we could observe is that since the Plywood Board was larger than the Acrylic one, the optimization algorithm chose a different grasping pose in Fig. \ref{fig:all_results}(e.2) in comparison with Fig. \ref{fig:all_results}(a.2) to reduce robotic workload. The reason was that the similar grasping pose was no longer reachable by the right robot arm.

The detailed parameters and costs of building the MSG used for both the two boards with different initial states are shown in Table \ref{tbl:details}. As can be seen from the table, the most time-consuming part was building the DCSG-RV (the $^\Theta$DCSG-RV$^1$ and $^\Theta$DCSG-RV$^2$ columns). The process involved repeated IK-solving, collision detection, and optimization. In the worst case, they cost more than an hour (the PB2 task). The finally merged MSG has, on average, 10,000 nodes and 100,000 edges (the MSG Info. column). 

\begin{table*}[!htbp]
    \caption{Detailed Parameters of the Manipulation State Graph}
    \centering
    \begin{threeparttable}
    \begin{tabular}{lllllllllllllll}
        \toprule
         & \multicolumn{2}{c}{${}^\#$Grasps} & & \multicolumn{2}{c}{DCSG Info.} & & \multicolumn{5}{c}{$^{\Theta}$DCSG-RV$^2$} & \multicolumn{2}{c}{MSG Info.}\\
         \cmidrule(lr){2-3} \cmidrule(lr){5-6} \cmidrule(lr){8-12} \cmidrule(lr){13-14}
         Name & Grip & Push & ${}^\#$Nodes & ${}^\#$Edges &  $^\Theta$DCSG & $^{\Theta}$DCSG-RV$^1$ & min & max & mean & std. & total  & ${}^\#$Nodes & ${}^\#$Edges & $^\Theta$MSG \\ \midrule
         AB1 & 30 & 64 & 151 & 2676 & 0.09  & 498.1  & 0.10 & 3.84 & 1.51 & 0.76 & 2983.4 & 9583  & 126279 & 351.6 \\
         AB2 & 30 & 64 & 151 & 2676 & 0.094 & 499.9  & 0.10 & 3.62 & 0.83 & 0.69 & 1642.7 & 7323  & 85476  & 351.6 \\
         AB3 & 30 & 64 & 151 & 2676 & 0.087 & 534.1  & 0.10 & 4.22 & 0.97 & 0.80 & 1912.1 & 6129  & 71954  & 340.6 \\
         PB1 & 36 & 72 & 151 & 2676 & 0.09  & 665.1 & 0.11 & 4.52 & 0.81 & 0.63 & 2143.1 & 21552 & 263039 & 628.6 \\
         PB2 & 36 & 72 & 151 & 2676 & 0.092 & 671.9 & 0.11 & 4.83 & 1.67 & 0.91 & 4421.1 & 10116 & 96763  & 654.7 \\
         \bottomrule
    \end{tabular}
    \begin{tablenotes}[flushleft]
    \item (Notes) ${}^\#$NAME: $^{\varnothing}$Number of NAME; $^\Theta$NAME: Time cost of NAME. The values are measured in seconds; $^\Theta$DCSG-RV$^1$: Time cost of geometric feasibility examination when building the DCSG-RVs. It includes the time of iterative IK-solving and collision detection; $^\Theta$DCSG-RV$^2$: Time cost of physical feasibility examination when building the DCSG-RVs.
    \end{tablenotes}
    \end{threeparttable}
    \label{tbl:details}
\end{table*}

The costs of graph search and detailed information of the final planned path are shown in Table \ref{tbl:searchdetails}. The search involved multiple starts and goals (the $^\#$Nodes columns). The worst search time was around 30 seconds (the PB1 task). The average length of a found path was around 20 (the Length column). There were 3-4 contact state changes along the found paths (the States in the parenthesis of the Length column). The maximum forces for the robot to manipulate objects when following the found paths were 12.7N for the acrylic board and 13.55N for the plywood board (the Max $F$ column).

\begin{table}[!htbp]
    \caption{Detailed Graph Search Results}
    \centering
    \begin{threeparttable}
    \begin{tabular}{llllll}
        \toprule
         & \multicolumn{2}{c}{${}^\#$Nodes} & \multicolumn{3}{c}{Path Info.} \\
         \cmidrule(lr){2-3} \cmidrule(lr){4-6}
         Name & ${}^\#$Start & ${}^\#$Goal & $^\Theta$Search & Length (States) & Max $F$ (N)\\ \midrule
         AB1 & 8  & 105 & 19.87 & 22 (3) & 12.70 \\
         AB2 & 8  & 99  & 18.63 & 20 (3) & 12.70 \\
         AB3 & 7  & 54  & 4.31  & 18 (4) & 12.70 \\
         PB1 & 10 & 40  & 32.35 & 19 (3) & 13.55 \\
         PB2 & 6  & 141 & 11.42 & 21 (3) & 13.55 \\
         \bottomrule
    \end{tabular}
    \begin{tablenotes}[flushleft]
    \item (Notes) ${}^\#$NAME: $^{\varnothing}$Number of NAME; $^\Theta$NAME: Time cost of NAME. The values are measured in seconds; Length (States): Length of found paths, with times of contact states changes in the parenthesis; Max $F$ (N): Maximum force for the robot to manipulate objects when following the found paths.
    \end{tablenotes}
    \end{threeparttable}
    \label{tbl:searchdetails}
\end{table}

\begin{figure*}[!htbp]
    \centering
    \includegraphics[width=\linewidth]{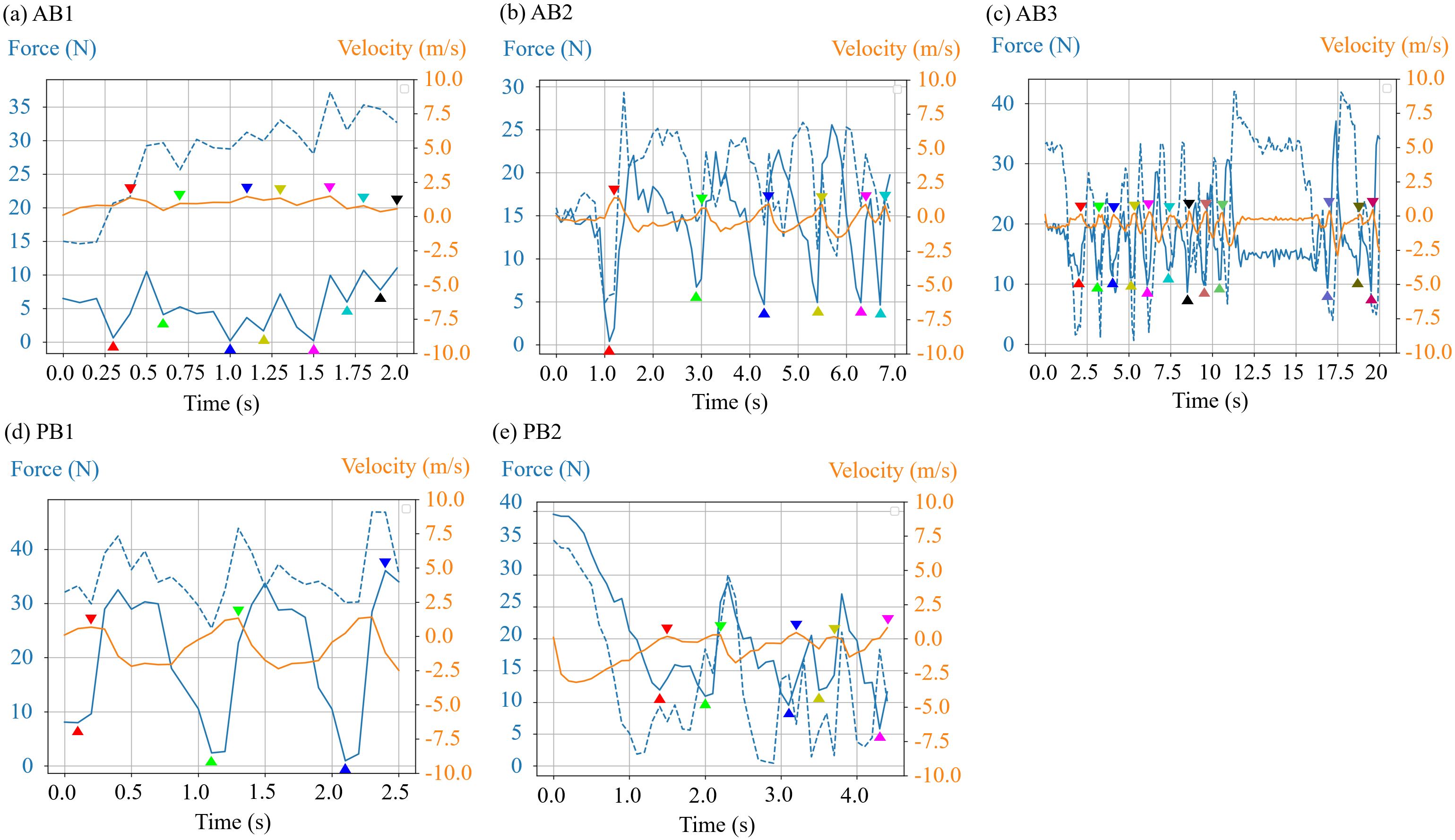}
    \caption{Changes of robot forces during the velocity-based lifter-follower control. Abbreviations: AB, Acrylic Board; PB, Plywood Board; Curves with different colors: Solid Blue, Vertical force of the pushing hand; Dashed Blue: Vertical force of the gripping hand; Orange: Output velocity of the velocity-based leader-follower controller. Triangle pairs with the same color highlight the drop-to-increase relation between forces of the pushing hand and the controller's output velocity.}
    \label{fig:curves}
\end{figure*}

The changes of robot forces during the velocity-based leader-follower control used in the second stage are shown in Fig. \ref{fig:curves}, with (a-c) displaying the force changes of the Acrylic Board task, (d,e) displaying the force changes of the Plywood Board task. In all cases, we asked a human to operate the switch of the vacuum lifter. The motion of the vacuum lifter and the boards were thus uneven and unstable. The different time lengths in the horizontal axes of Fig. \ref{fig:curves} reveal the unevenness. The randomly changing force curves (solid and dashed blue curves) reveal the instability of human operations. On the robot side, we controlled the two arms by requesting them to follow the pushing hand since, at the end of the manipulation stage, the robot always used both arms to keep the objects in non-contact states. 

The solid blue curves in Fig. \ref{fig:curves} show the pushing hand's force changes in the vertical direction. The orange curves are the output speed of the controller. By comparing the two curves, we could see that the output speed increased when the forces of the pushing hand dropped. The triangle pairs with the same colors around the curves highlight the related drop-to-increase correspondence. Also, by comparing the two curves, we could see that the speed decreased when the forces increased (not highlighted). The robot hands move faster or slower according to the force changes\footnote{Although the theoretical velocity could be negative, we did not allow the actual hand motion to be reversed in the implementation. The hands' motion was stopped when the output speed was lower than zero.}. 

The dashed blue curves in Fig. \ref{fig:curves} are the gripping hand's force changes. As expected, the gripping hand's forces were less sensitive than the pushing hand. The changes of the blue dashed curves were, in most cases, slower than the blue curves. An exception happened at the beginning of Fig. \ref{fig:curves}(e), where the gripping hand's force dropped earlier than the pushing hands. However, despite the earlier drop, the pushing hand's force also dropped quickly. It motivated a positive speed (see the red arrow on the orange curve) and avoided the overload emergency failure. 

The force curves confirmed that using the velocity output computed based on the pushing hand's force changes could successfully coordinate the gripping hand's movement while maintaining balance with the vacuum lifter and the pushing hand.

\begin{figure*}[!htbp]
    \centering
    \includegraphics[width=\linewidth]{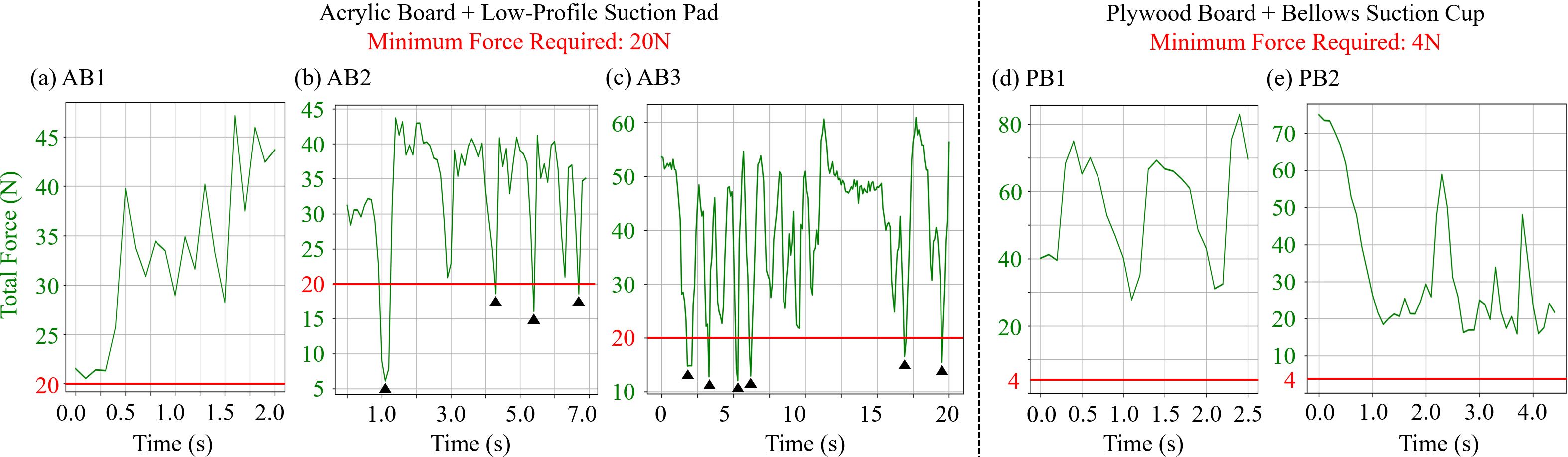}
    \caption{Sum of the forces provided by the two hands during lift-up. Abbreviations: AB, Acrylic Board; PB, Plywood Board; Green Curve: Sum of the forces; Red lines: Minimum forces required for the vacuum lifter's suction force to hold plates in the air. The minimum force is the difference between the weight of a plate and the maximum suction force.}
    \label{fig:fsum}
\end{figure*}

Fig. \ref{fig:fsum} shows the total forces provided by the two hands during the lift-up. The curves in the respective subfigures are the sum of the solid and dashed blue curves in the correspondent subfigures of Fig. \ref{fig:curves}. The red horizontal line indicates the minimum force that needs to be provided. If the total force from the hands is smaller than this minimum value, the vacuum lifter will bear a force larger than the maximum vacuum force of the suction cup and thus will fail to hold the plate in the air continuously. By observing the results, we could see that the total forces are larger than the minimum value in most cases. The observation demonstrates that the real-world data are coherent with the theoretical optimization results. On the other hand, there are several short exceptions labeled by the black triangles. The provided forces dropped smaller than the minimum required forces in these cases, and the vacuum lifter's suction forces can no longer keep the plates. However, the drop was instant and dynamic. The controller responded quickly to speed up the robot hands and avoid failures.

\section{Conclusions and Future Work}
\label{sec:conclusions}

This paper presented a planning and control method for a dual-arm robot to raise a heavy plate with the help of a vacuum lifter. The heavy board was assumed to be initially placed on a table. In the planning part, a planner built and searched an MSG to find combined prehensile and non-prehensile manipulation while considering supporting forces from the table surface and suction forces provided by the vacuum lifter. In the control part, a velocity-based force controller was used to control the speed of robot arms to lift the board high to the air. Experimental results showed that the planner and controller could successfully find manipulation sequences and generate safe dual-arm motion to coordinate with a vacuum lifter operated by a human. With the planner and controller's help, a robot can work with a vacuum lifter to expand its manipulation ability.

In the future, we are interested in replacing the stationary dual-arm robots with mobile manipulators and developing practical systems to support logistics in a large warehouse.


\bibliographystyle{IEEEtran}
\bibliography{citations.bib}

\end{document}